\UseRawInputEncoding
\documentclass[lettersize,journal]{IEEEtran}
\usepackage{amsmath,amsfonts}
\usepackage{algorithmic}
\usepackage{array}
\usepackage[caption=false,font=normalsize,labelfont=sf,textfont=sf]{subfig}
\usepackage{textcomp}
\usepackage{stfloats}
\usepackage{url}
\usepackage{verbatim}
\usepackage{graphicx}
\def\BibTeX{{\rm B\kern-.05em{\sc i\kern-.025em b}\kern-.08em
    T\kern-.1667em\lower.7ex\hbox{E}\kern-.125emX}}
\usepackage{balance}

\usepackage{hyperref}

\usepackage{multirow}
\usepackage{booktabs}
\newcommand{\etal}{\emph{et al. }}

\newcommand{\ie}{\emph{i.e. }}

\usepackage{newtxmath}

\usepackage{colortbl}

\usepackage{algorithm}

\begin{document}
\title{Knowledge Transfer-Driven Few-Shot Class-Incremental Learning}
\author{Ye Wang, Yaxiong Wang, Guoshuai Zhao, and Xueming Qian, \textit{Member, IEEE}
\thanks{
This work was supported by in part by the NSFC under Grant 62272380.

Ye Wang (E-mail: {xjtu2wangye@stu.xjtu.edu.cn}), Yaxiong Wang (E-mail: {wangyx15@stu.xjtu.edu.cn}), and Guoshuai Zhao (E-mail: {guoshuai.zhao@xjtu.edu.cn}) are with the SMILES LAB at School of Information and Communications Engineering, Faculty of Electronic and Information Engineering, Xi’an Jiaotong University, Xi’an, 710049, China.



Xueming Qian (corresponding author, E-mail: {qianxm@mail.xjtu.edu.cn}) is with the Ministry of Education Key Laboratory for Intelligent Networks and Network Security and with SMILES LAB, Xi’an Jiaotong University, Xi’an, 710049, China. 

}}

\markboth{Journal of \LaTeX\ Class Files,~Vol.~18, No.~9, September~2020}%
{How to Use the IEEEtran \LaTeX \ Templates}

\maketitle

\begin{abstract}


Few-shot class-incremental learning (FSCIL) aims to continually learn new classes using a few samples while not forgetting the old classes.
The key of this task is effective knowledge transfer from the base session to the incremental sessions.
Despite the advance of existing FSCIL methods, the proposed knowledge transfer learning schemes are sub-optimal due to the insufficient optimization for the model's plasticity.
To address this issue, we propose a Random Episode Sampling and Augmentation (RESA) strategy that relies on diverse pseudo incremental tasks as agents to achieve the knowledge transfer.
Concretely, RESA mimics the real incremental setting and constructs pseudo incremental tasks globally and locally, where the global pseudo incremental tasks are designed to coincide with the learning objective of FSCIL and the local pseudo incremental tasks are designed to improve the model's plasticity, respectively.
Furthermore, to make convincing incremental predictions, we introduce a complementary model with a squared Euclidean-distance classifier as the auxiliary module, which couples with the widely used cosine classifier to form our whole architecture.
By such a way, equipped  with model decoupling strategy, we can maintain the model's stability while enhancing the model's plasticity.
Extensive quantitative and qualitative experiments on three popular FSCIL benchmark datasets demonstrate that our proposed method, named Knowledge Transfer-driven Relation Complementation Network (KT-RCNet), outperforms almost all prior methods.
More precisely, the average accuracy of our proposed KT-RCNet outperforms the second-best method by a margin of \textbf{5.26}$\%$, \textbf{3.49}$\%$, and \textbf{2.25}$\%$ on \textit{mini}ImageNet, CIFAR100, and CUB200, respectively.
Our code is available at \href{https://github.com/YeZiLaiXi/KT-RCNet.git }{https://github.com/YeZiLaiXi/KT-RCNet.git}.

\end{abstract}

\begin{IEEEkeywords}
lifelong learning, class-incremental learning, few-shot class-incremental learning, image recognition.
\end{IEEEkeywords}
\section{Introduction}\label{sec:intro}

\begin{figure}[ht]
\centering
\includegraphics[width=1.0\columnwidth]{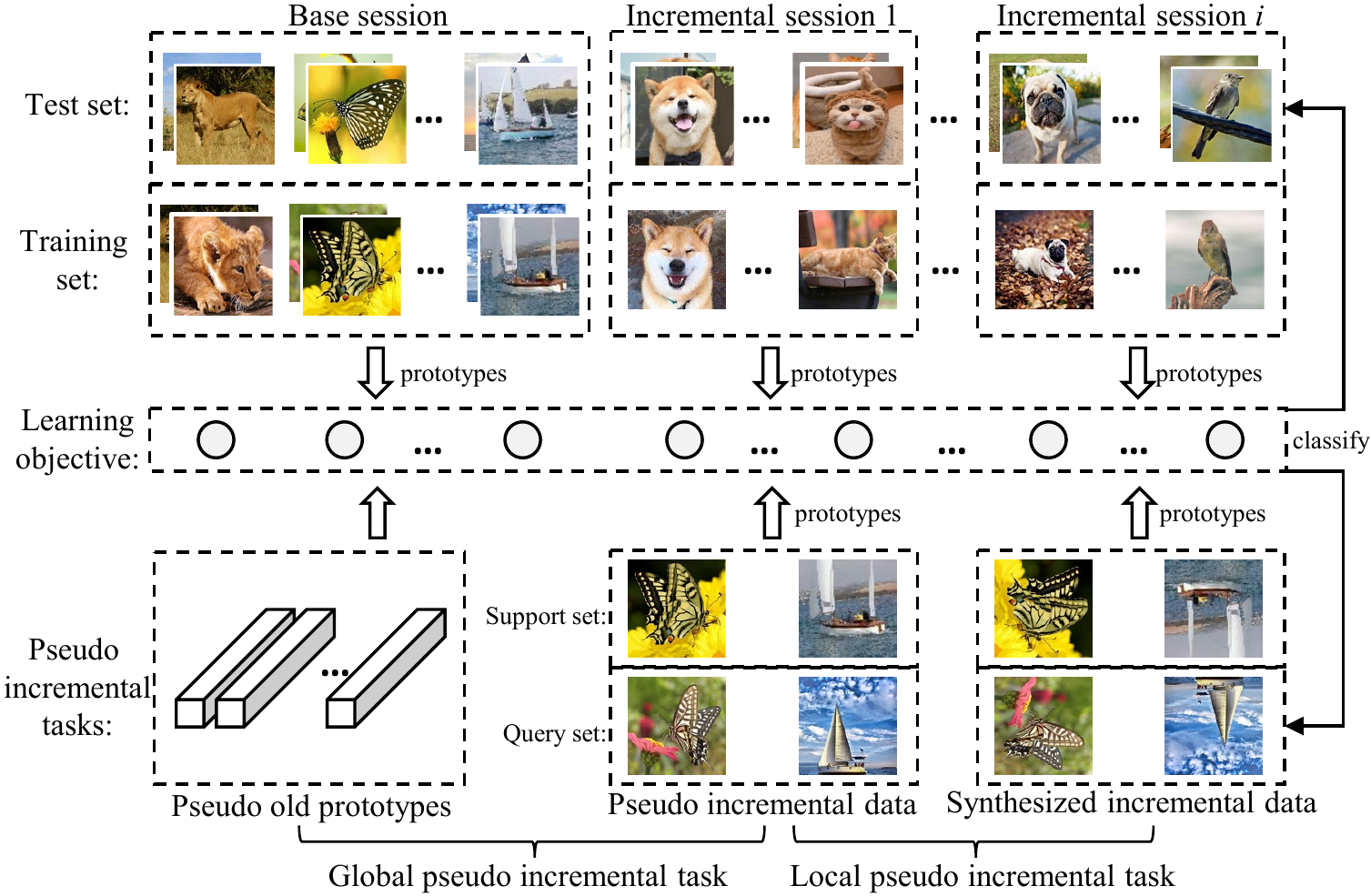}
\caption{The learning objective of FSCIL is to learn a classifier initialized by prototypes to classify the test sets of all encountered classes. Our proposed method mimics the real incremental setting and constructs pseudo incremental tasks from the global and local perspectives to coincide with the learning objective of FSCIL and improve the model's plasticity.
\label{motiv}}
\end{figure}

\IEEEPARstart{A}{l}though Deep Neural Networks have achieved great success in many vision tasks \cite{9810872, 9764821, li2021dbcface}, these methods can only process predefined classes.
In many real image recognition scenarios, the number of classes that needs to be processed usually grows continually.
The conventional solution referred to as joint training is to use the data of old and new classes to train the model.
Obviously, this strategy costs substantial time and effort to train the model in each learning phase. 
In response to this weakness, class-incremental learning (CIL) is proposed~\cite{lwf}, aiming to learn new classes fast while not forgetting the old classes. 
Despite the advance of current CIL methods \cite{9705128,wang2022memory}, the key factor behinds the success is the large amount of annotated training samples for new classes.
However, annotating a large number of training samples still costs time and effort, and the number of training samples in some scenarios, such as identifying rare bird species, is scarce, which often makes these methods fail due to the overfitting problem.
In response to such challenging incremental scenarios, few-shot class-incremental learning (FSCIL)~\cite{Tao_2020_CVPR} is proposed to learn new classes using a few samples while keeping the performance on the old classes.

FSCIL inherits the characteristics of CIL and few-shot learning (FSL): several learning sessions come in sequence like the common CIL, but the samples in each new class are limited, as FSL assumes.
The first session, dubbed the base session, provides sufficient training samples for model learning.
In contrast, the following sessions, called incremental sessions, only possess limited training samples.
In each session, the model is trained only with the current session's data but evaluated on the test sets of all encountered classes.
The scarcity of new training samples will seriously destroy the model's stability and plasticity.

To mitigate these problems, researchers propose to the model decoupling strategy that freezes the encoder in incremental sessions to maintain the model's stability~\cite{Zhu_2021_CVPR,Zhang_2021_CVPR,hersche2022constrained}.
However, this strategy also results in the model relying heavily on the knowledge learned in the base session for recognizing the incremental classes.
To improve the model's plasticity, previous methods construct various pseudo incremental tasks with the data sampled from the base session aiming to transfer the knowledge learned from the base session to the incremental sessions.
Despite the advance of these methods, the constructed pseudo incremental tasks are either local classification tasks that do not align with the learning objective of FSCIL~\cite{Zhang_2021_CVPR,Chi_2022_CVPR}, or global classification tasks that have limited effect on improving the model's plasticity~\cite{Zhu_2021_CVPR}.
Specifically, the local classification task only includes pseudo new classes, while FSCIL aims to learn a global classifier capable of classifying both old and new classes. This inconsistency between the local classification task and the learning objective of FSCIL inevitably affects the model's performance. On the other hand, the global classification task, which consists of pseudo old and new classes, matches the learning objective of FSCIL but only consists of data sampled from the base session, thereby compromising its effectiveness in enhancing the model's plasticity.
By borrowing the treasure from the few-shot learning~\cite{NIPS2016_90e13578}, one straightforward approach is to utilize the pseudo new classes to construct few-shot-based local classification tasks to enhance the model's plasticity. However, a delicately pretrained model can classify these pseudo new classes well, the contribution of these local classification tasks is compromised. 
Based on above considerations, we propose a knowledge transfer learning scheme called Random Episode Sampling and Augmentation (RESA).
Our proposed RESA scheme focuses on constructing pseudo incremental tasks specifically designed to improve the model's plasticity. 
As illustrated in Figure \ref{motiv}, each pseudo incremental task consists of three components: the pseudo old prototypes, the pseudo incremental data, and the synthesized incremental data. 
By combining the pseudo old prototypes and the pseudo incremental data, we create a series of global pseudo incremental tasks to optimize the model, ensuring that the learning objective is consistent with that of FSCIL.
We further enhance the diversity of pseudo new classes by introducing the synthesized incremental data. 
This allows us to construct more effective local pseudo incremental tasks by combining the pseudo incremental data with the synthesized incremental data, resulting in further improvements to the model's plasticity.

Furthermore, on the design of robust metric criteria, researchers pay little attention to this study, and the problem is rarely explored. 
Previous methods~\cite{Zhu_2021_CVPR,Zhang_2021_CVPR,hersche2022constrained,Chi_2022_CVPR,Zhou_2022_CVPR} mainly rely on a single metric to estimate the relation between prototypes and test features. 
We think such a mechanism is unilateral. 
The reason mainly stems from the poor representation of the new classes, with the unreliable representation, the model is easy to give improper relation prediction.
Intuitively, with only one metric, no mechanism to calibrate it if the captured relation is inappropriate.  
To remedy this issue, we further augment our system with dual-metric learning, whose target is to jointly estimate the relevance of the prototype and the test features.
To be specific, we introduce a complementary model with a squared Euclidean-distance classifier as the auxiliary module, which couples with the widely used cosine classifier to form our full architecture referred to as  Knowledge Transfer-driven Relation Complementation Network (KT-RCNet). 
Two classifiers provide dual metrics to give a more convincing relation estimation.

Our main contributions in this paper are summarized as follows:
\begin{itemize}
    \item \textbf{A knowledge transfer learning} scheme, RESA, is specially designed for FSCIL, in which we construct pseudo incremental tasks from global and local perspectives to help the model transfer the knowledge learned from the base session to incremental sessions.
    \item \textbf{A relation complementation strategy} is proposed, which ensembles different metrics to investigate the comprehensive relation of prototypes and test features.
    \item \textbf{Competitive performance}. Extensive experiments on \textit{mini}ImageNet, CIFAR100, and CUB200 datasets demonstrate the superiority of our proposed method over previous methods. 
\end{itemize}

\section{Related Work}\label{sec:related}

\subsection{Few-Shot Learning}
Few-shot learning (FSL) aims to develop machine learning algorithms capable of processing new classes using only a few samples.
Scarce training samples provide limited prior knowledge, making it challenging for the model to recognize new classes.
To address this issue, the learning paradigm in FSL is often organized as meta-tasks similar to the inference task. 
Based on this learning paradigm, many interesting works are proposed, which can be categorized into metric-based, optimization-based, and hallucination-based methods.
The metric-based methods\cite{NIPS2017_cb8da676,Chikontwe_2022_CVPR,Liu_2022_CVPR} leverage different metrics or networks, such as Euclidean or Graph Neural Networks(GNNs), to construct the nearest neighbor classifier to measure the similarity between the support and query samples.
The optimization-based methods\cite{pmlr-v70-finn17a,rusu2018metalearning,baik2020learning,fei2021melr} design different meta-learners or optimization strategies to learn to adapt to different query set with the support set, 
For example, Finn \etal \cite{pmlr-v70-finn17a} propose a classical and famous method named MAML, consisting of a meta-learner using the support set and a fixed learning rate to conduct model fast adaption.
Rusu \etal \cite{rusu2018metalearning} decouple the gradient-based adaptation procedure from the underlying high-dimensional space of model parameters to a low-dimensional space to make the model generalize to new tasks easier. 
Baik \etal \cite{baik2020learning} propose task-and-layer-wise attenuation on the compromised initialization to reduce the adverse effects of forcibly sharing the initialization in MAML. 
The hallucination-based methods focus on learning a generation model or module to generate classification weights \cite{Guo_2020_CVPR} or fake samples \cite{Dong_2022_CVPR,Xu_2022_CVPR}.
For example, Dong \etal \cite{Guo_2020_CVPR} propose a method that utilizes the attention mechanism and fuses the information provided by both support and query set to generate the classification weights to classify query samples.
Xu \etal \cite{Xu_2022_CVPR} propose a method that uses the conditional variational autoencoder to generate more representative features, while Dong \etal \cite{Dong_2022_CVPR} propose to generate the adversarial images to improve the representation ability of the model.
Though this research field is similar to FSCIL, most FSL methods do not consider the performance on old classes, while FSCIL aims to achieve good performance on both old and new classes.

\subsection{Class-Incremental Learning}
Class-incremental learning (CIL) aims to enable continuous learning of new classes while retaining knowledge of previously learned classes. 
However, due to the limitation of using old data, the model's parameters are overwritten by the data of new classes in incremental sessions, leading to the notorious catastrophic forgetting problem.
To address this issue, current CIL methods can be roughly categorized into four groups: regularization-based, rehearsal-based, isolation-based, and rehearsal-free approaches.
The regularization-based methods\cite{lwf,PODNet,Hu_2021_CVPR} distill the knowledge from previous tasks when training the new task to prevent the model from forgetting old, such as Li \etal \cite{lwf} propose to distill the outputs of classification model while Douillard \etal \cite{PODNet} distill the features of each stage in the feature extraction process. 
The rehearsal-based methods \cite{Rebuffi_2017_CVPR,Yan_2022_CVPR,wang2022memory} adopt different strategies, such as reservoir sampling \cite{NEURIPS2019_fa7cdfad}, to restore samples of the previous task and then use them either as inputs or constrain to alleviate catastrophic forgetting when train the new task.
To store more old samples with limited memory, Wang \etal \cite{wang2022memory} propose to reduce the image's quality.
The isolation-based methods \cite{Yan_2021_CVPR,Liu_2021_CVPR} introduce extra parameters for each new task, such as Yan \etal \cite{Yan_2021_CVPR} train new feature encoder for each new task.
With the emerging of foundation models\cite{radford2021learning,dosovitskiy2020image}, the rehearsal-free methods design various prompt-based strategies to learn corresponding knowledge for different incremental tasks~\cite{wang2022learning,wang2022sprompts}.
Despite the advance of current CIL methods, these methods often assume there are sufficient training samples for new class learning, which are not suitable for some incremental scenarios where training samples for new classes are limited.

\begin{figure*}[ht]
\centering
\includegraphics[width=2.0\columnwidth]{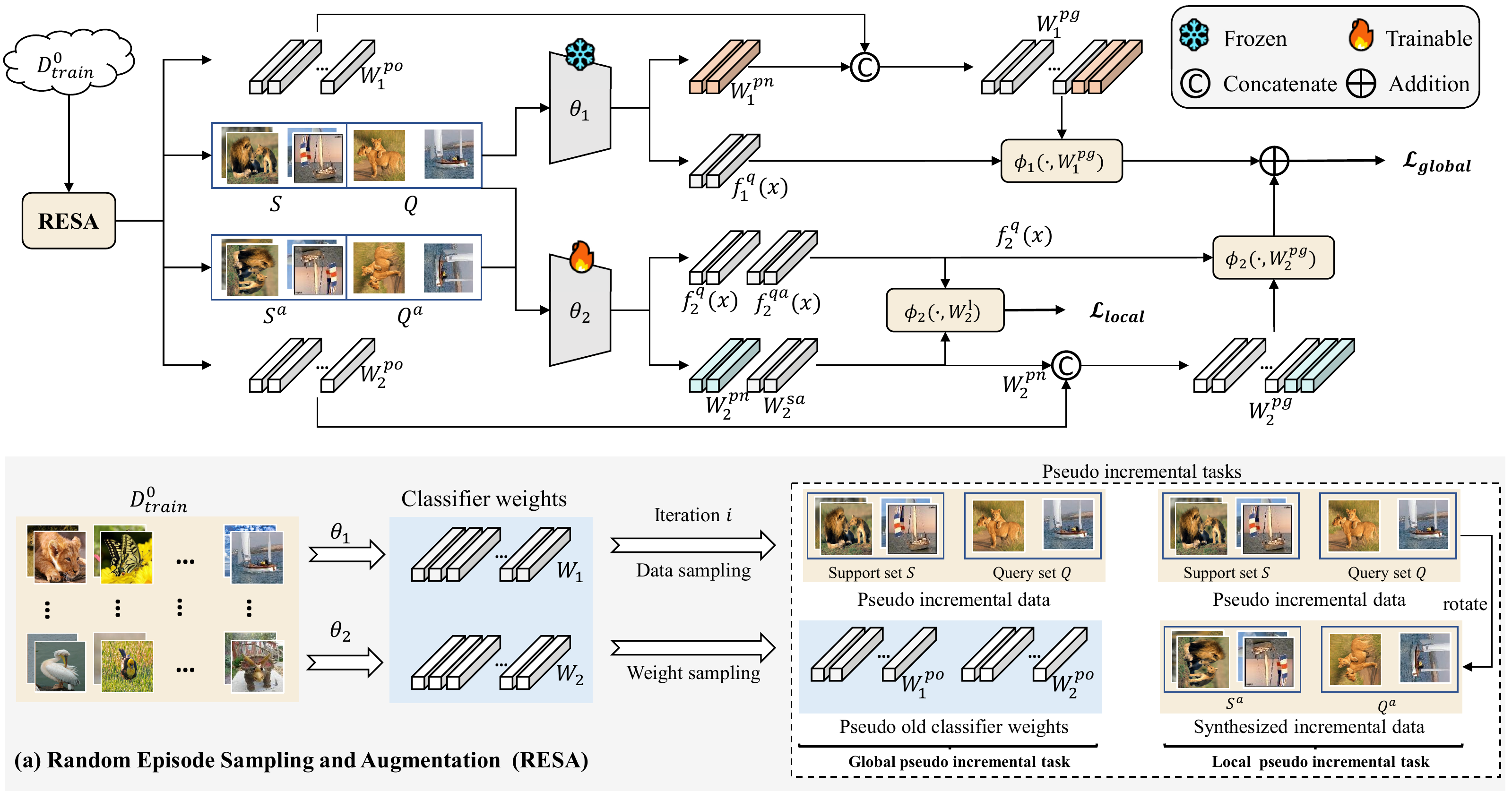}
\caption{ The overview of our proposed knowledge transfer learning scheme, where (a) the random episode sampling and augmentation strategy.
\label{method}}
\end{figure*}

\subsection{Few-Shot Class-Incremental Learning}
FSCIL aims to learn a global classifier in phases, where the number of training samples provided for new class learning is scarce. 
Due to this task's challenging and practical character, it has attracted the attention of many scholars in recent years.
To maintain the model's stability, the knowledge distillation strategy is adopted by most FSCIL methods \cite{Dong_Hong_Tao_Chang_Wei_Gong_2021,Cheraghian_2021_CVPR,10138923}.
For example, Dong \etal \cite{Dong_Hong_Tao_Chang_Wei_Gong_2021} propose to distill the relation between different classes to balance the tasks of old-knowledge preserving and new-knowledge adaptation.
Cheraghian \etal \cite{Cheraghian_2021_CVPR} propose a semantic guide method that distills the semantic information which is aligned by an attention module to prevent catastrophic forgetting.
In contrast to knowledge distillation-based methods, many methods propose to freeze the encoder in the incremental sessions and demonstrate the effectiveness of such a strategy~\cite{shi2021overcoming,Zhang_2021_CVPR,hersche2022constrained}.
However, this strategy also constrain the model's plasticity.
To solve this issue, mimicking the incremental setting to construct pseudo incremental tasks becomes an emerging and effective solution.
For example, 
Zhu \etal \cite{Zhu_2021_CVPR} propose a strategy named Random Episode Selection Strategy (RESS) that samples part of old data as pseudo new data and the prototypes of other old classes as the pseudo old prototypes to construct pseudo incremental tasks.
Zhang \etal \cite{Zhang_2021_CVPR} propose a Pseudo Incremental Learning (PIL) strategy that construct the pseudo incremental tasks by sampling part of old data as pseudo old data and rotating the sampled data to synthesize pseudo new data.
Chi \etal \cite{Chi_2022_CVPR} propose a meta-learning strategy that mimics the multi-step incremental setting and constructs sequential pseudo incremental tasks to make the model learn to optimize itself using a few training samples. 
The essential mechanism of RESS is to construct the global pseudo incremental tasks to optimize the model's global relation prediction ability.
However, the effectiveness of this strategy is limited due to the fact that a pretrained model can already classify old classes well.
The essential mechanism of strategies proposed by Zhang \etal \cite{Zhang_2021_CVPR} and Chi \etal \cite{Chi_2022_CVPR} is to construct local pseudo incremental tasks to optimize the model's plasticity.
However, these strategies are inconsistent with learning objective of FSCIL which compromises the model's global relation prediction ability.
Recently, considering that single model mainly focuses on one-side knowledge which limits the ability of resisting catastrophic forgetting, Ji \etal \cite{MCNet} propose to ensemble different model to capture diverse knowledge to mitigate such limitation, where a CNN architecture is introduced to capture global knowledge and a Transformer architecture is introduced to capture local knowledge.
Different from previous methods, our proposed strategy constructs the pseudo incremental tasks globally and locally, and ensemble different metrics to give convincing incremental relation estimations rather than different architectures.

\section{Preliminary Knowledge}\label{sec:problem}
Before delving into the details of our methodology, we first introduce the problem definition of few-shot class-incremental learning (FSCIL). 
FSCIL aims to learn a global classifier in phases to classify all encountered classes and proposes an incremental setting as follow.
It's worth noting that each learning phase is also called the session in FSCIL.
Formally, let $\mathcal{D}^0\rightarrow{\mathcal{D}^1}\rightarrow {...}$ denote the data stream. 
The classes contained in different sessions satisfy ${\mathcal{C}^{i}}\cap{\mathcal{C}^{j}}=\emptyset(i\neq{j})$.
Each $\mathcal{D}^i$ consists of a training set $\mathcal{D}^{i}_{train}$ and a test set $\mathcal{D}^{i}_{test}$, where only $\mathcal{D}^0_{train}$ contains lots of samples, $\mathcal{D}^{i}_{train}(i>0)$ contains a few samples, such as 5 training samples for each class.
In session $i$, only $\mathcal{D}^{i}_{train}$ is available.
In contrast, $\left\{{\mathcal{D}^{0}_{test},...,\mathcal{D}^{i}_{test}}\right\}$ are used to evaluate the model's performance.
Under the background of model decoupling strategy, the essential problem that needs to be solved in FSCIL is an incremental relation measuring problem between the classifier weights initialized by prototypes of training data and the test features.
However, scarce training samples make such a problem challenging in incremental sessions.
\section{Method}\label{sec:method}
In this section, we first describe the overall framework in Section \ref{sec:framework}. 
Then, we describe the conventional training paradigm in Section\ref{sec:std}.
Next, we detail how to apply the random episode sampling and augmentation (RESA) strategy to constructed pseudo incremental tasks to optimize the model in Section \ref{sec:resa} and \ref{sec:cl}, respectively.
Finally, we detail the inference in Section \ref{sec:inf}. 

\subsection{Framework Overview}
\label{sec:framework}
Our proposed method consists of a base model with the cosine classifier as previous FSCIL methods~\cite{Zhu_2021_CVPR,Zhang_2021_CVPR,kang2023on} and a complementary model with the squared Euclidean-based classifier.
We first adopt the conventional training paradigm to learn the parameters of the base model.
Then, as shown in Figure \ref{method}, we utilize the RESA to construct pseudo incremental tasks to learn the parameters of the complementary model.
In the end, we use prototypes to initialize or expand the classifiers of the base model and the complementary model to prepare for future incremental relation measuring.
For the sake of following description,  we refer the following feature encoding of the base model and the complementary model as $f_1(x) = \mathcal{N}(x; {\theta}_1)$ and $f_2(x) = \mathcal{N}(x; {\theta}_2)$, where ${\theta}_1$ and ${\theta}_2$ refer to the parameter of the base model's and the complementary model's encoder.


\vspace{0.2cm}
\subsection{Conventional training paradigm}
\label{sec:std}
Sufficient training samples in the base session enable us to train a satisfactory classification model to classify base classes.
However, if simply employing a linear layer as the classification layer, this will results the imbalance magnitude between base and future coming incremental classes \cite{hou2019learning,Zhang_2021_CVPR}, compromising the model's performance.
Therefore, we replace the linear classification layer with the cosine classifier.
Concretely, let $x$ denote the image data. We first input the $x$ to the base model and compute the classification score $P$ as follow:
\begin{equation}
\label{eq:cos}
    P = \text{softmax}(s{\Phi}_1(f_1(x), W_1)),
\end{equation}
where $s$ is the scale factor, ${\Phi}_1(a, b)=\frac{a\cdot{b}}{||a||_2{||b||_2}}$ is the cosine classifier, $\cdot$ refers to the inner product, and $W_1$ refers to the classifier weights.
After obtained the classification score $P$, the parameters ${\theta}_1$ and $W_1$ is optimized by
\begin{equation}
    {\theta}_1^*, W_1^*=\mathop{\arg\min}\limits_{{\theta}_1, W_1}\mathcal{L}_{ce}(P, y),
\end{equation}
where $\mathcal{L}_{ce}$ refers to the cross-entropy loss function, $y$ is the ground truth of $x$.

\vspace{0.2cm}
\subsection{Random Episode Sampling and Augmentation}
\label{sec:resa}
In FSCIL, the scarce training samples in the incremental sessions make it is difficult to further train the model, resulting in a common problem, \ie, the representations for new classes are weak.
Due to the weak representations for new classes, the model is easy to give improper relation prediction.
For this problem, our solution is to use a complementary model with different metric from the base model and ensemble different metrics to mitigate this problem.
To learn the parameters of the complementary model, a straightforward method is to train it in the conventional manner.
However, the data forms in the incremental sessions are few-shot based, the task gap between the base session and the incremental sessions makes such a strategy sub-optimal.
To well transfer the knowledge learned from the base session to the incremental sessions, RESA mimics the real incremental setting to construct a series of pseudo incremental tasks from global and local perspectives for each episode with the data sampled from the base session, where the global pseudo incremental task is used to coincide with the learning objective of FSCIL, and the local pseudo incremental task is used to improve the model's plasticity.
To construct these tasks, there are four main steps, weight computing, data sampling, weight sampling, and data augmentation.

\vspace{0.2cm}
$\bullet$ \textbf{Weight computing.}
To prepare for later weight sampling, with $D_{train}^0$, RESA first get the classifier weights $W_1$ of the base model by 
\begin{equation}
\label{eq:mu1}
    W_1 = \text{mean}(f_1(x))\in{\mathbb{R}^{N\times{d}}},
\end{equation}
where $N$ refers to the number of base classes, and $d$ refers to the dimension of data embedding.
Next, RESA applies the complementary model to encode $D_{train}^0$.
In the end, the classifier weights $W_2$ of the complementary model are computed  by
\begin{equation}
\label{eq:mu2}
    W_2 = \text{mean}(f_2(x))\in{\mathbb{R}^{N\times{d}}},
\end{equation}
where $f_2(x)$ denotes the data embedding of $D_{train}^0$ encoded by the complementary model.

\vspace{0.2cm}
$\bullet$ \textbf{Data sampling.}
To mimic the data setting of new coming classes in the incremental session, RESA first randomly selects several classes from ${\mathcal{C}^{0}}$ as the pseudo incremental classes.
Then, RESA randomly samples a few data for each selected class to constitute the support set $S$ and query set $Q$, where $S$, $Q$ will serve as the training, test set of real incremental classes, respectively.

\vspace{0.2cm}
$\bullet$ \textbf{Prototype sampling.}
To mimic the data setting of old classes in the incremental session, except the classifier weights of pseudo incremental classes, other classifier weights of $W_1$ and $W_2$ are selected as the pseudo old classifier weights ${W}_1^{po}$ and  ${W}_2^{po}$ of the base model and complementary model, respectively.

\vspace{0.2cm}
$\bullet$ \textbf{Data augmentation.}
To improve the model's plasticity, RESA performs data augmentation to synthesize pseudo incremental data to enhance the diversity of pseudo new classes.
Concretely, RESA rotates the $S$ and $Q$ to construct the support set $S^a$ and query set $Q^a$ of new pseudo incremental classes.

Overall, the combination of $\left\{S, Q, S^a, Q^a, {W}_1^{po}, {W}_2^{po}\right\}$ forms a pseudo incremental task, where the combination of $\left\{S, Q, {W}_1^{po}, {W}_2^{po}\right\}$ forms the global pseudo incremental task, and the combination of $\left\{S, Q, S^a, Q^a\right\}$ forms the local pseudo incremental task.

\begin{algorithm}[ht]
\caption{Complementary learning.}
  \label{alg2}
  \begin{algorithmic}[1]
  \REQUIRE
     The base model $f(;{\theta}_1)$,  the complementary model $f(;{\theta}_2)$, global pseudo incremental task $\left\{S, Q, S^a, Q^a, {W}_1^{po}, {W}_2^{po}\right\}$, local pseudo incremental task $\left\{S, Q, S^a, Q^a\right\}$.
  \ENSURE
    A trained $f(;{\theta}_2)$.
   \WHILE{not done}
        \STATE ${W}_1^{pn},{W}_2^{pn} \gets$ Get the base model's and complementary model's pseudo new classifier weights using $S$, Eq. \ref{eq:mu1} and Eq. \ref{eq:mu2}, respectively.
        \STATE ${W}_1^{pg} \gets$ Get the base model's pseudo global classifier weights by concatenating ${W}_1^{pn}$ and ${W}_1^{po}$
        \STATE ${W}_2^{pg} \gets$ Get the complementary model's pseudo global classifier weights by concatenating ${W}_2^{pn}$ and ${W}_2^{po}$
        \STATE $P_{global} \gets$ Make predictions for $Q$ using ${W}_1^{pg}, {W}_2^{pg}$, Eq. \ref{relation:r1}, \ref{relation:r2} and \ref{pre:p_global}
        \STATE $\mathcal{L}_{global} \gets$ Compute the global loss by using Eq.\ref{loss:global}
        \STATE ${W}_2^{sa} \gets$ Get the synthesized pseudo new classifier weights using $S^a$ and Eq. \ref{eq:mu2}
        \STATE ${W}_2^{l} \gets$ Get the local classifier weights by concatenating ${W}_2^{pn}$ and ${W}_2^{sa}$
        \STATE $P_{local} \gets$ Make predictions for $\left\{Q, Q^a\right\}$$ $ using ${W}_2^{l}$, Eq. \ref{relation:r2} and softmax function with a scale factor
        \STATE $\mathcal{L}_{local} \gets$ Compute the local loss using $P_{local}$ and Eq. \ref{loss:local}
        \STATE $\mathcal{L} \gets$ Compute the total loss using Eq.\ref{loss:all}
        \STATE Optimize the complementary model with SGD
        
   \ENDWHILE
  \end{algorithmic}
\end{algorithm}

\vspace{0.2cm}
\subsection{Complementary Learning}
\label{sec:cl}
With the constructed pseudo incremental tasks, the complementary learning aims to optimize the complementary model for relation calibration. 
The pseudo code of this stage is illustrated in Alg.\ref{alg2}.
With the constructed global pseudo incremental task $\left\{S, Q, {W}_1^{po}, {W}_2^{po}\right\}$,
we first encode the data of $S$ and $Q$ using $f(;{\theta}_1)$ and $f(;{\theta}_2)$.
The corresponding data embeddings are denoted as $f_1^s(x)$, $f_1^q(x)$, $f_2^s(x)$, and $f_2^q(x)$, respectively.
Then, the pseudo new classifier weights  ${W}_1^{pn}$ of the base model are computed using Eq.\ref{eq:mu1} and $f_1^s(x)$.
Analogously, the pseudo new classifier weights ${W}_2^{pn}$ of the complementary model are computed using Eq.\ref{eq:mu2} and $f_2^s(x)$.
Next, ${W}_1^{pn}$ and ${W}_1^{po}$ are concatenated as the pseudo global classifier weights ${W}_1^{pg}$ of the base model, while  ${W}_2^{pn}$ and ${W}_2^{po}$ are concatenated as the pseudo global classifier weights ${W}_2^{pg}$ of the complementary model.
Given the pseudo global classifier weights ${W}_1^{pg}$ and ${W}_2^{pg}$, the respective relations can be calculated accordingly:
\begin{align}
    \label{relation:r1}
    r_1 &= {\Phi}_1(f_1^q(x), {W}_1^{pg}),\\
    \label{relation:r2}
    r_2 &= {\Phi}_2(f_2^q(x), {W}_2^{pg}),
\end{align}
where $r_1$, $r_2$ are the relation estimations given by the base model and complementary model,respectively, ${\Phi}_2(a, b)=-||a-b||^2/d$
refers to the squared Euclidean distance-based classifier.
With $r_1$ and $r_2$,  the final incremental relation measuring $P_{global}$ is given by integrating the above two predictions:
\begin{equation}
\label{pre:p_global}
    P_{global} = \text{softmax}(s(\frac{r_1}{d}+r_2)),
\end{equation}
where $d$ is used to eliminate the impact of dimension.
Finally, the global loss $\mathcal{L}_{global}$ is computed by the cross entropy (CE) loss:
\begin{equation}
\label{loss:global}
    \mathcal{L}_{global} = \mathcal{L}_{CE}({P_{global}}, Y_{global}),
\end{equation}
where $Y_{global}$ refers to the ground truth of the query data contained in the global pseudo incremental task.

With the constructed local pseudo incremental task $\left\{S, Q, S^a, Q^a\right\}$, 
We first encode the data of $S^a$ and $Q^a$ using the $f(;{\theta}_2)$.
The corresponding data embeddings of $S^a$ and $Q^a$ are denoted as $f_2^{sa}(x)$ and $f_2^{qa}(x)$, respectively.
Then, we use Eq. \ref{eq:mu2} to compute the mean features ${W}_2^{sa}$  of $f_2^{sa}(x)$.
Next, we concatenate  ${W}_2^{sa}$ and ${W}_2^{pn}$ as the local classifier weights ${W}_2^{l}$ to classify $f_2^{qa}(x)$ and $f_2^q(x)$  using  Eq. \ref{relation:r2}.
Let the computed relation be $r_{local}$, the local relation estimation $P_{local}$ is predicted 
\begin{equation}
    P_{local} =  \text{softmax}(sr_{local}).
\end{equation}
Consequently, the local loss is defined as:
\begin{equation}
\label{loss:local}
    \mathcal{L}_{local} = \mathcal{L}_{CE}(P_{local}, Y_{local}),
\end{equation}
where $Y_{local}$ refers to the ground truth of the query data contained in the local pseudo incremental task. 

To coincide with the learning objective of FSCIL and improve the model's new class adaption ability, we define the  total objective as:
\begin{equation}
\label{loss:all}
    \mathcal{L} = {\lambda}_1\mathcal{L}_{global} + {\lambda}_2\mathcal{L}_{local},
\end{equation}
where ${\lambda}_1$ and ${\lambda}_2$ are hyper-parameters to balance the two losses.

\subsection{Incremental Relation Measuring}
\label{sec:inf}
In incremental sessions, the training set of current incremental sessions is first used to expand the previous classifiers.
For example, let $W_1^o$ and $W_2^o$ denote the old classifier weights of the base model and complementary model.
In each incremental session, we first get the data embedding of the available training set by inputting the training set to the base model and complementary model, respectively.
Then, we use Eq.~\ref{eq:mu1} and Eq.~\ref{eq:mu2} to compute the classifier weights $W_1^n$ and $W_2^n$ of new classes, respectively.
Next, we use the concatenation of $W_1^o$ and  $W_1^n$ to expand the old classifier of the base model.
The update classifier weights we denote as $W_1^g$.
Similarly, the old classifier weights $W_2^o$ of the complementary model is expanded by the concatenation of $W_2^o$ and  $W_2^n$.
The update classifier weights of the complementary model we denote as $W_2^g$.
Given a test sample $x$ from the test set of all encountered classes, the incremental relation estimation $P$ is given by
\begin{equation}
    P = {\Phi}_1(f_1(x), W_1^g) + {\Phi}_2(f_2(x), W_2^g).
\end{equation}
\section{Experiments}\label{sec:exp}

\begin{table*}[ht]\small
\begin{center}
\caption{Comparison with other methods on \textit{mini}ImageNet, where * represents the results copied from \cite{Tao_2020_CVPR}.}
\label{tab:cmp_sota_mini}
\begin{tabular}{l|ccccccccc|cc}
\toprule
\multirow{2}{*}{Method} & \multicolumn{9}{c|}{sessions} & \multirow{2}{*}{\texttt{Avg.}} & \multirow{2}{*}{\texttt{Diff.}} \\
\cmidrule{2-10}
{} & 0 & 1 & 2 & 3 & 4 & 5 & 6 & 7 & 8 & {} & {} \\
\midrule
\midrule
Joint-CNN                               & 81.20	& 75.62	& 70.66	& 65.81	& 62.20	& 58.41	& 55.78	& 53.16	& 50.00 & {63.65} & {0.00} \\
NCM$^{*}$ \cite{Hou_2019_CVPR}          & 61.31 & 47.80 & 39.31 & 31.91 & 25.68 & 21.35 & 18.67 & 17.24 & 14.17 & 30.83   & -35.83 \\
iCaRL$^{*}$ \cite{Rebuffi_2017_CVPR}    & 61.31 & 46.32 & 42.94 & 37.63 & 30.49 & 24.00 & 20.89 & 18.80 & 17.21 & 33.29   & -32.79  \\
EEIL$^{*}$ \cite{Castro_2018_ECCV}      & 61.31 & 46.58 & 44.00 & 37.29 & 33.14 & 27.12 & 24.10 & 21.57 & 19.58 & 34.97   & -30.42 \\
TOPIC\cite{Tao_2020_CVPR}               & 61.31 & 50.09 & 45.17 & 41.16 & 37.48 & 35.52 & 32.19 & 29.46 & 24.42 & {39.64} & {-25.58 } \\
SPPR\cite{Zhu_2021_CVPR}                & 61.45 & 63.80 & 59.53 & 55.53 & 52.50 & 49.60 & 46.69 & 43.79 & 41.92 & {52.76} & {-8.08 } \\
CEC\cite{Zhang_2021_CVPR}               & 72.00 & 66.83	& 62.97	& 59.43	& 56.70	& 53.73	& 51.19	& 49.24	& 47.63 & {57.75} & {-2.37} \\
F2M\cite{shi2021overcoming}             & 72.05 & 67.47 & 63.16 & 59.70 & 56.71 & 53.77 & 51.11 & 49.21 & 47.84 & {57.89} & {-2.16 }\\
MCNet\cite{MCNet}                       & 72.33 & 67.70 & 63.50 & 60.34 & 57.59 & 54.70	& 52.13	& 50.41	& 49.08	& {58.64} & {-0.92} \\
MetaFSCIL\cite{Chi_2022_CVPR}           & 72.04 & 67.94 & 63.77 & 60.29 & 57.58 & 55.16 & 52.90 & 50.79 & 49.19 & {58.85} & {-0.81 } \\
FACT\cite{Zhou_2022_CVPR}               & 72.56 & 69.63 & 66.38 & 62.77 & 60.60 & 57.33 & 54.34 & 52.16 & 50.49 & {60.70} & {+0.49 }\\
C-FSCIL\cite{hersche2022constrained}    & 76.40	& 71.14	& 66.46	& 63.29	& 60.42	& 57.46	& 54.78	& 53.11	& 51.41 & {61.61} & {+1.41 } \\
SoftNet\cite{kang2023on}                & 79.77 & 75.08 & 70.59 & 66.93 & 64.00 & 61.00 & 57.81 & 55.81 & 54.68 & {65.07} & {+4.68 } \\
\textbf{KT-RCNet(Ours)}                    & \textbf{84.62} & \textbf{79.94} & \textbf{75.70} & \textbf{72.21} & \textbf{69.38} & \textbf{66.26} & \textbf{63.48} & \textbf{61.39} & \textbf{60.02} & \textbf{70.33} & \textbf{+10.02} \\

\bottomrule
\end{tabular}
\end{center} 
\end{table*}

\begin{table*}[ht]\small
\begin{center}
\caption{Comparison with other methods on CIFAR100, where * represents the results copied from \cite{Tao_2020_CVPR}.} 
\label{tab:cmp_sota_cifar}
\begin{tabular}{l|ccccccccc|cc}
\toprule
\multirow{2}{*}{Method} & \multicolumn{9}{c|}{sessions} & \multirow{2}{*}{\texttt{Avg.}} & \multirow{2}{*}{\texttt{Diff.}}\\
\cmidrule{2-10}
{} & 0 & 1 & 2 & 3 & 4 & 5 & 6 & 7 & 8 & {} & {} \\
\midrule
\midrule
Joint-CNN                               & 80.15	& 74.57	& 69.93	& 65.31	& 61.00	& 57.79	& 54.47	& 51.59	& 49.66 & {62.72} & {0.00}\\
NCM$^{*}$ \cite{Hou_2019_CVPR}          & 64.10 & 53.05 & 43.96 & 36.97 & 31.61 & 26.73 & 21.23 & 16.78 & 13.54 & {34.22} & {-36.12 }\\
iCaRL$^{*}$ \cite{Rebuffi_2017_CVPR}    & 64.10 & 53.28 & 41.69 & 34.13 & 27.93 & 25.06 & 20.41 & 15.48 & 13.73 & {32.87} & {-35.93 }\\
EEIL$^{*}$ \cite{Castro_2018_ECCV}      & 64.10 & 53.11 & 43.71 & 35.15 & 28.96 & 24.98 & 21.01 & 17.26 & 15.85 & {33.79} & {-33.81 }\\
TOPIC\cite{Tao_2020_CVPR}               & 64.10 & 55.88 & 47.07 & 45.16 & 40.11 & 36.38 & 33.96 & 31.55 & 29.37 & {42.62} & {-20.29 }\\
SPPR\cite{Zhu_2021_CVPR}                & 63.97 & 65.86 & 61.31 & 57.60 & 53.39 & 50.93 & 48.27 & 45.36 & 43.32 & {54.45} & {-6.34 }\\
CEC\cite{Zhang_2021_CVPR}               & 73.07	& 68.88	& 65.26	& 61.19	& 58.09	& 55.57	& 53.22	& 51.34	& 49.14 & {59.53} & {-0.52 }\\
F2M\cite{shi2021overcoming}             & 71.45 & 68.10 & 64.43 & 60.80 & 57.76 & 55.26 & 53.53 & 51.57 & 49.35 & {59.14} & {-0.31 }\\
MetaFSCIL\cite{Chi_2022_CVPR}           & 74.50 & 70.10 & 66.84 & 62.77 & 59.48 & 56.52 & 54.36 & 52.56 & 49.97 & {60.79} & {+0.31 }\\
C-FSCIL\cite{hersche2022constrained}    & 77.47 & 72.40 & 67.47 & 63.25 & 59.84 & 56.95 & 54.42 & 52.47 & 50.47 & {61.64} & {+0.81 }\\
MCNet\cite{MCNet}                       & 73.30 & 69.34 & 65.72 & 61.70 & 58.75 & 56.44	& 54.59	& 53.01	& 50.72	 & {60.40} & {+1.06 }\\
FACT\cite{Zhou_2022_CVPR}               & 74.60 & 72.09 & 67.56 & 63.52 & 61.38 & 58.36 & 56.28 & 54.24 & 52.10 & {62.24} & {+2.44 }\\
SoftNet\cite{kang2023on}                & 79.88 & 75.54 & 71.64 & 67.47 & 64.45 & 61.09 & 59.07 & 57.29 & 55.33 & {65.75} & {+5.67 }\\
\textbf{KT-RCNet(Ours)}                           & \textbf{83.40 } & \textbf{78.75} & \textbf{74.94} & \textbf{70.81} & \textbf{67.84} & \textbf{64.89} & \textbf{63.10} & \textbf{60.92} & \textbf{58.53} & \textbf{69.24} & \textbf{+8.87 }\\

\bottomrule
\end{tabular}
\end{center}
\end{table*}

\begin{table*}[ht]\small
\begin{center}
\caption{Comparison with other methods on CUB200, where * represents the results copied from \cite{Tao_2020_CVPR}.} 
\label{tab:cmp_sota_cub}
\begin{tabular}{l|ccccccccccc|cc}
\toprule
\multirow{2}{*}{Method} & \multicolumn{11}{c|}{sessions} & \multirow{2}{*}{\texttt{Avg.}} & \multirow{2}{*}{\texttt{Diff.}}\\
\cmidrule{2-12}
{} & 0 & 1 & 2 & 3 & 4 & 5 & 6 & 7 & 8 & 9 & 10 & {} & {}\\
\midrule
\midrule
Joint-CNN                               & 78.68	& 73.49	& 69.86	& 66.10	& 64.74	& 62.47	& 60.64	& 59.32	& 57.25	& 57.67	& 57.50 & {64.34} & {0.00}\\
NCM$^{*}$ \cite{Hou_2019_CVPR}          & 68.68 & 57.12 & 44.21 & 28.78 & 26.71 & 25.66 & 24.62 & 21.52 & 20.12 & 20.06 & 19.87 & 32.49 & -37.63  \\
iCaRL$^{*}$ \cite{Rebuffi_2017_CVPR}    & 68.68 & 52.65 & 48.61 & 44.16 & 36.62 & 29.52 & 27.83 & 26.26 & 24.01 & 23.89 & 21.16 & 36.67 & -36.34 \\
EEIL$^{*}$ \cite{Castro_2018_ECCV}      & 68.68 & 53.63 & 47.91 & 44.20 & 36.30 & 27.46 & 25.93 & 24.70 & 23.95 & 24.13 & 22.11 & 36.27 & -35.39 \\
TOPIC\cite{Tao_2020_CVPR}               & 68.68 & 62.49 & 54.81 & 49.99 & 45.25 & 41.40 & 38.35 & 35.36 & 32.22 & 28.31 & 26.28 & {43.92} & {-31.22}\\
SPPR\cite{Zhu_2021_CVPR}                & 68.68 & 61.85 & 57.43 & 52.68 & 50.19 & 46.88 & 44.65 & 43.07 & 40.17 & 39.63 & 37.33 & {49.34} & {-20.17}\\
CEC\cite{Zhang_2021_CVPR}               & 75.85	& 71.94	& 68.50	& 63.50	& 62.43	& 58.27	& 57.73	& 55.81	& 54.83	& 53.52	& 52.28 & {61.33} & {-5.22 }\\
MetaFSCIL\cite{Chi_2022_CVPR}           & 75.90 & 72.41 & 68.78 & 64.78 & 62.96 & 59.99 & 58.30 & 56.85 & 54.78 & 53.82 & 52.64 & {61.93} & {-4.86 }\\
F2M\cite{shi2021overcoming}             & 77.13 & 73.92 & 70.27 & 66.37 & 64.34 & 61.69 & 60.52 & 59.38 & 57.15 & 56.94 & 55.89 & {63.96} & {-1.61 }\\
SoftNet\cite{kang2023on}                & 78.07 & 74.58 & 71.37 & 67.54 & 65.37 & 62.60 & 61.07 & 59.37 & 57.53 & 57.21 & 56.75 & {64.68} & {-0.75 }\\
FACT\cite{Zhou_2022_CVPR}               & 75.90 & 73.23 & 70.84 & 66.13 & 65.56 & 62.15 & 61.74 & 59.83 & 58.41 & 57.89 & 56.94 & {64.42} & {-0.56 }\\
MCNet\cite{MCNet}                       & 77.57 & 73.96 & 70.47 & 65.81 & 66.16 & 63.81	& 62.09	& 61.82	& 60.41	& 60.09	& 59.08 & {65.57} & {+1.58 }\\
\textbf{KT-RCNet(Ours)} & \textbf{79.86}	& \textbf{76.48} & \textbf{73.34} & \textbf{69.72} & \textbf{68.48} & \textbf{65.93} & \textbf{64.58} & \textbf{63.68} & \textbf{62.04} & \textbf{61.48} & \textbf{60.47} & \textbf{67.82} & \textbf{+2.97}\\

\bottomrule 
\end{tabular}

\end{center}
\end{table*}

\subsection{Datasets}
\noindent\textbf{\textit{mini}ImageNet.} \textit{mini}ImageNet is the subset of ImageNet \cite{2015ImageNet} dataset, which comprises 100 classes. Each class consists of 500 training images and 100 test images. Following the approach proposed in \cite{Tao_2020_CVPR}, we split this dataset into 60 base classes and 40 incremental classes. The 40 incremental classes are further divided equally into 8 incremental sessions, where each incremental session takes the setting of 5-way-5-shot. This indicates that each session consists of 5 classes, and each class has five training images. 

\vspace{0.2cm}
\noindent\textbf{CIFAR100.} CIFAR100 \cite{cifar} consists of 60,000 RGB images from 100 classes, where each class consists of 500 training images and 100 test images. Following \cite{Tao_2020_CVPR}, we split this dataset into 60 base classes and 40 incremental classes. The 40 incremental classes are further equally divided into 8 incremental sessions, where each session takes the setting of 5-way-5-shot.

\vspace{0.2cm}
\noindent\textbf{Caltech-UCSD Birds-200-2011.} CUB200 \cite{cub} is a fine-grained dataset that contains 11,788 RGB images from 200 classes, where each class consists of approximately 30 training images and 30 test images. Following \cite{Tao_2020_CVPR}, we split this dataset into 100 base classes and 100 incremental classes. The 100 incremental classes are equally divided into 10 incremental sessions, where each session takes the setting of 10-way-5-shot.

\subsection{Implementation Details}
Our implementation is based on the PyTorch \cite{paszke2019pytorch} platform. Like \cite{Zhu_2021_CVPR, shi2021overcoming}, we adopt ResNet18 as the backbone for all benchmark datasets. 

$\bullet$ In the pretraining stage, the base model is trained for 50 epochs with a batch size of 128 on CUB200 using the SGD optimizer with a learning rate of 0.03, weight decay of 0.0001 and momentum of 0.9. The learning rate is decreased by a factor of 0.1 per 10 epochs.
For \textit{mini}ImageNet and CIFAR100, we set the batch size to 64 and the learning rate and weight decay to 0.1 and 0.0005, respectively. The learning rate is decreased by a factor of 0.1 every 40 epochs.

$\bullet$ In the knowledge transfer learning stage, we train the complementary model for 80 epochs. In each epoch, we randomly sample 200 tasks, and we use the SGD optimizer with a learning rate of 0.03, weight decay of 0.0001, and momentum of 0.9. The learning rate is decreased by a factor of 0.1 per 20 epochs.
The scale factors are set to 16, 16, 12 for CIFAR100, CUB200, and \textit{mini}ImageNet respectively.
Random resized crop, random horizontal flip, and color jitter techniques are employed for data augmentation during training, following Zhu \etal \cite{Zhu_2021_CVPR}.

\subsection{Evaluation Protocol}
In the inference stage of each session, we use the test sets of all encountered classes to evaluate the model's performance and report the Top-1 accuracy.
To evaluate the model's overall performance, we compute the average accuracy \texttt{Avg.}$=\frac{1}{M+1}\sum_{i=0}^{M}\mathcal{A}_{i}$  across all sessions, where $M$ represents the number of incremental sessions and $\mathcal{A}_{i}$ represents the Top-1 accuracy of the $i-$th session.
Following previous class incremental learning methods \cite{wang2022learning,wang2022dualprompt}, we also compute the performance gap \texttt{Diff.}$=\mathcal{A}_{M}-\mathcal{A}_M^{ub}$ between the method and the upper-bound method Joint-CNN, where Joint-CNN represents the method that uses both the training data of old and new classes to train the model in each session, and $\mathcal{A}_M^{ub}$ represents the accuracy of the last session of Joint-CNN.

\begin{table*}[ht]\small
\begin{center}
\caption{Ablation studies on \textit{min}ImageNet. \textbf{RESA} refers to the random episode sampling and augmentation. Compared with single metric, ensembling different metrics achieves better performance, and our proposed RESA can further boost the performance.} 
\begin{tabular}{ccc|ccccccccc}
\toprule
\multirow{2}{*}{B-model} & \multirow{2}{*}{C-model} & \multirow{2}{*}{RESA} & \multicolumn{9}{c}{sessions} \\
\cmidrule{4-12}
{} & {} & {} & 0 & 1 & 2 & 3 & 4 & 5 & 6 & 7 & 8 \\
\midrule
\midrule
$\surd$ & {} & {}                   & {81.87} & {76.82} & {72.29} & {68.60} & {65.44} & {62.45} & {59.57} & {57.42} & {55.98}  \\
{} & {$\surd$} & {}                 & {80.83} & {74.32} & {70.20} & {66.68} & {63.79} & {60.92} & {58.34} & {56.61} & {55.18}  \\
$\surd$ & {$\surd$} & {}            & {82.98} & {77.60} & {73.31} & {69.87} & {67.03} & {64.15} & {61.56} & {59.83} & {58.70}  \\
\midrule
{} & {$\surd$} & {$\surd$}          & {82.80} & {77.66} & {73.59} & {69.83} & {67.00} & {63.48} & {60.49} & {58.38} & {56.68}  \\
{$\surd$} & {$\surd$} & {$\surd$}   & \textbf{84.62} & \textbf{79.94} & \textbf{75.70} & \textbf{72.21} & \textbf{69.38} & \textbf{66.26} & \textbf{63.48} & \textbf{61.39} & \textbf{60.02}  \\
\bottomrule
\end{tabular}
\end{center}
\label{tab:ablat}
\end{table*}

\subsection{Quantitative Comparison}
To validate the effectiveness of our proposed method, we compare it with some classical class-incremental learning methods (iCaRL~\cite{Rebuffi_2017_CVPR}, EEIL~\cite{Castro_2018_ECCV}, and NCM~\cite{Hou_2019_CVPR}) and recent FSCIL methods (TOPIC \cite{Tao_2020_CVPR}, SPPR \cite{Zhu_2021_CVPR}, CEC \cite{Zhang_2021_CVPR}, F2M \cite{shi2021overcoming}, C-FSCIL \cite{hersche2022constrained}, MetaFSCIL \cite{Chi_2022_CVPR}, FACT \cite{Zhou_2022_CVPR}, SoftNet~\cite{kang2023on} and MCNet~\cite{MCNet}) on three popular benchmark datasets.
As can be observed from Table \ref{tab:cmp_sota_mini}, \ref{tab:cmp_sota_cifar}, and \ref{tab:cmp_sota_cub}, 
\begin{itemize}
    \item On three benchmark datasets, due to scarce training samples in the incremental sessions, class incremental learning methods iCaRL, EEIL and NCM overfit the training data, resulting in significant performance degradation as the learning process proceeds.
    \item On \textit{mini}ImageNet, our proposed method achieves the highest accuracy on each session. Particularly, compared with the second-best method SoftNet, the average accuracy \texttt{Avg.} of our proposed KR-RCNet has an improvement of \textbf{5.26}$\%$, while the performance gap \texttt{Diff.} with the Joint-CNN of our proposed KR-RCNet has an improvement of \textbf{5.34}$\%$.
    \item On CIFAR100, our proposed KR-RCNet also achieves the highest accuracy on each session. Particularly, compared with the second-best method SoftNet, the average accuracy \texttt{Avg.} of our proposed KR-RCNet has an improvement of \textbf{3.49}$\%$, while the performance gap \texttt{Diff.} with the Joint-CNN of our proposed KR-RCNet has an improvement of \textbf{3.2}$\%$.
    \item On CUB200, our proposed KR-RCNet achieves the best performance on each session as on \textit{mini}ImageNet and CIFAR100. Particularly, compared with the second-best method MCNet, the average accuracy \texttt{Avg.} of our proposed KR-RCNet has an improvement of \textbf{2.25}$\%$, while the performance gap \texttt{Diff.} with the Joint-CNN of our proposed KR-RCNet has an improvement of \textbf{1.39}$\%$.
\end{itemize}
In summary, compared to other methods shown in Table \ref{tab:cmp_sota_mini}, \ref{tab:cmp_sota_cifar}, and \ref{tab:cmp_sota_cub}, our proposed method not only achieves the highest average accuracy \texttt{Avg.} but also achieves the largest performance gap \texttt{Diff.} with the Joint-CNN. The quantitative comparisons demonstrate the superior performance of our proposed KT-RCNet.

\subsection{Ablation Study}
To validate the effectiveness of each component in RCNet, we conduct several ablation studies on \textit{mini}ImageNet.
As we can see from Table \ref{tab:ablat}, the accuracy on the last session given by only using the base model (B-model) is 55.98$\%$ (row 1) while that given by only using the complementary model (C-model) is 55.18$\%$ (row 2).
When using the C-model as an auxiliary model to complement the B-model (row 3), the accuracy on the last session is 58.70$\%$ which surpasses that given by using the B-model or the C-model alone by a margin of \textbf{2.72}$\%$ and \textbf{3.52}$\%$, respectively.
The results demonstrate that relying on a single metric to perform incremental relation prediction is insufficient, and ensembling different metrics is an effective strategy for FSCIL.
When using RESA to train the C-model (row 4), we can see that our proposed RESA boosts the C-Model (row 2) performance from 55.18$\%$ to 56.68$\%$.
When using RESA to train the C-model and using C-model to complement the B-model (row 5), we can see that our proposed RESA boosts the performance of the full model (row 3) from 58.70$\%$ to 60.02$\%$.
The results validate the effectiveness of our proposed RESA.

\begin{table*}[ht]\small
\begin{center}
\caption{Further Analysis on Random Episode Sampling and Augmentation (RESA), where RESS \cite{Zhu_2021_CVPR}, PIL\cite{Zhang_2021_CVPR}, and Meta-learning\cite{Chi_2022_CVPR} are the previously proposed knowledge transfer learning strategies, C refers to CutMix \cite{yun2019cutmix} , M refers to MixUp\cite{zhang2017mixup}, and R refers to Rotate\cite{Zhang_2021_CVPR}.} 
\begin{tabular}{lccccccccccc}
\toprule
\multirow{2}{*}{Strategy}  & \multicolumn{11}{c}{sessions} \\
\cmidrule{2-12}
{} & 0 & 1 & 2 & 3 & 4 & 5 & 6 & 7 & 8 & 9 & 10\\
\midrule
\midrule
RESS\cite{Zhu_2021_CVPR}             & {78.21} & {73.81} & {70.78} & {67.22} & {66.16} & {63.82} & {62.54} & {61.29} & {59.68} & {58.92} & {58.10} \\
PIL\cite{Zhang_2021_CVPR}            & 79.35 & 76.15 & 72.64 & 68.59 & 67.44 & 65.07 & 63.72 & 62.72 & 60.87 & 60.20 & 59.18 \\
Meta-learning\cite{Chi_2022_CVPR}    & 79.54 & 76.18 & 72.51 & 68.58 & 67.14 & 64.63 & 63.32 & 62.38 & 60.75 & 60.20 & 59.32 \\
RESA-C (Ours)  & {79.01} & {75.47} & {72.15} & {68.31} & {67.21} & {64.96} & {63.89} & {63.06} & {61.33} & {60.83} & {59.85} \\
RESA-M (Ours)  & {79.48} & {76.35} & {73.03} & {69.30} & {68.19} & {65.70} & {64.28} & {63.67} & {61.81} & {61.22} & {59.99} \\
RESA-R (Ours)  & \textbf{79.86}	& \textbf{76.48} & \textbf{73.34} & \textbf{69.72} & \textbf{68.48} & \textbf{65.93} & \textbf{64.58} & \textbf{63.68} & \textbf{62.04} & \textbf{61.48} & \textbf{60.47} \\
\bottomrule
\end{tabular}

\end{center}
\label{tab:resa}
\end{table*}

\begin{figure*}[ht]
\centering
\includegraphics[width=2.0\columnwidth]{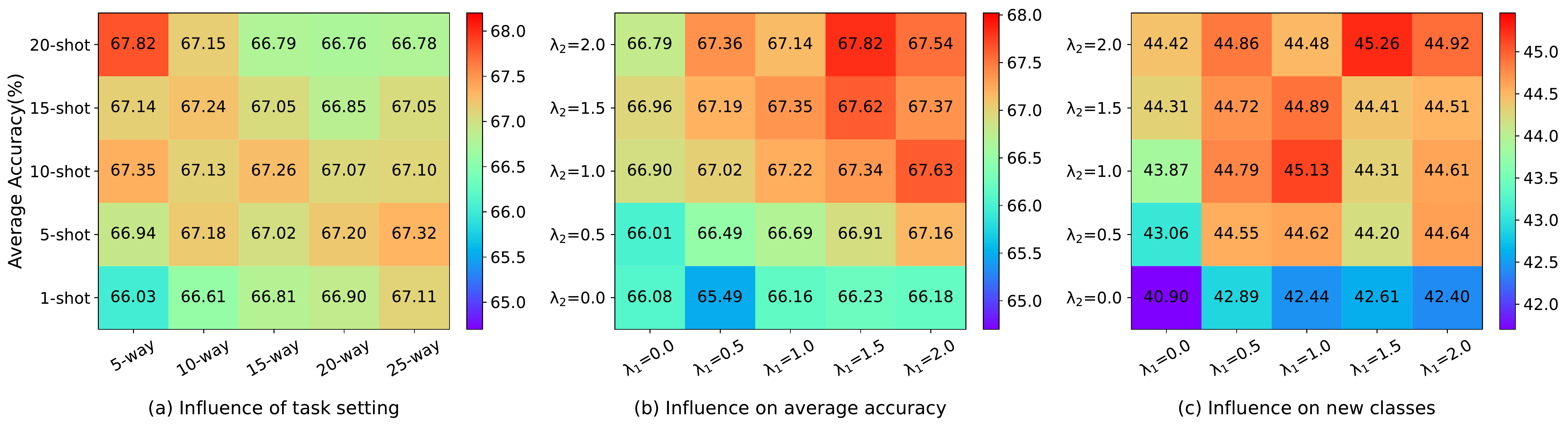}
\caption{Performance analysis under different conditions on CUB200, where (a) we use different $N$-way-$K$-shot settings to sample the data of pseudo incremental classes, we change ${\lambda}_1$ and ${\lambda}_2$ among different values and report the (b) average accuracy and (c) the accuracy on new classes.
\label{fig:hyper}}
\end{figure*}

\subsection{Discussion}

\vspace{0.2cm}
\subsubsection{The impact of sampling setting}
In RESA, we adopt $N$-way-$K$-shot setting to sample the data of pseudo incremental classes from the base session.
To study the impact of the $N$-way-$K$-shot setting on average accuracy in the knowledge transfer learning stage, we change the number of ways among $\left\{5, 10, 15, 20, 25\right\}$ and the number of shots among $\left\{1, 5, 10, 15, 20\right\}$.
The results given by different combinations are reported in Figure \ref{fig:hyper}(a). No matter fixing the number of ways or shots, our proposed method can achieve a satisfactory result as long as suitable shots or ways are set.
Particularly, setting the number of ways to 5 and the number of shots to 20 achieves the best performance on CUB200.
The main reason we guess may be that the data of constructed pseudo tasks is sampled from the base session.
Setting a small way can reduce overfitting, and a large shot can provide sufficient prior information.

\vspace{0.2cm}
\subsubsection{The impact of $\mathcal{L}_{global}$ and $\mathcal{L}_{local}$.}
To verify the effectiveness of $\mathcal{L}_{global}$ and $\mathcal{L}_{local}$, we change ${\lambda}_1$ and ${\lambda}_2$ among $\left\{0, 0.5, 1, 1.5, 2.0\right\}$ and report the average accuracy and the accuracy on new classes.
As shown in Figure \ref{fig:hyper}(b), $\mathcal{L}_{global}$ and $\mathcal{L}_{local}$  together are better than using $\mathcal{L}_{global}$ or $\mathcal{L}_{local}$ alone.
What's more, it's clear from Figure \ref{fig:hyper}(c) that the $\mathcal{L}_{local}$ is beneficial to improve the model's plasticity.
Particularly, setting ${\lambda}_1$ to 1.5 and ${\lambda}_2$ to 2.0 is an optimal configuration.

\begin{figure}[h]
\centering
\includegraphics[width=1.0\columnwidth]{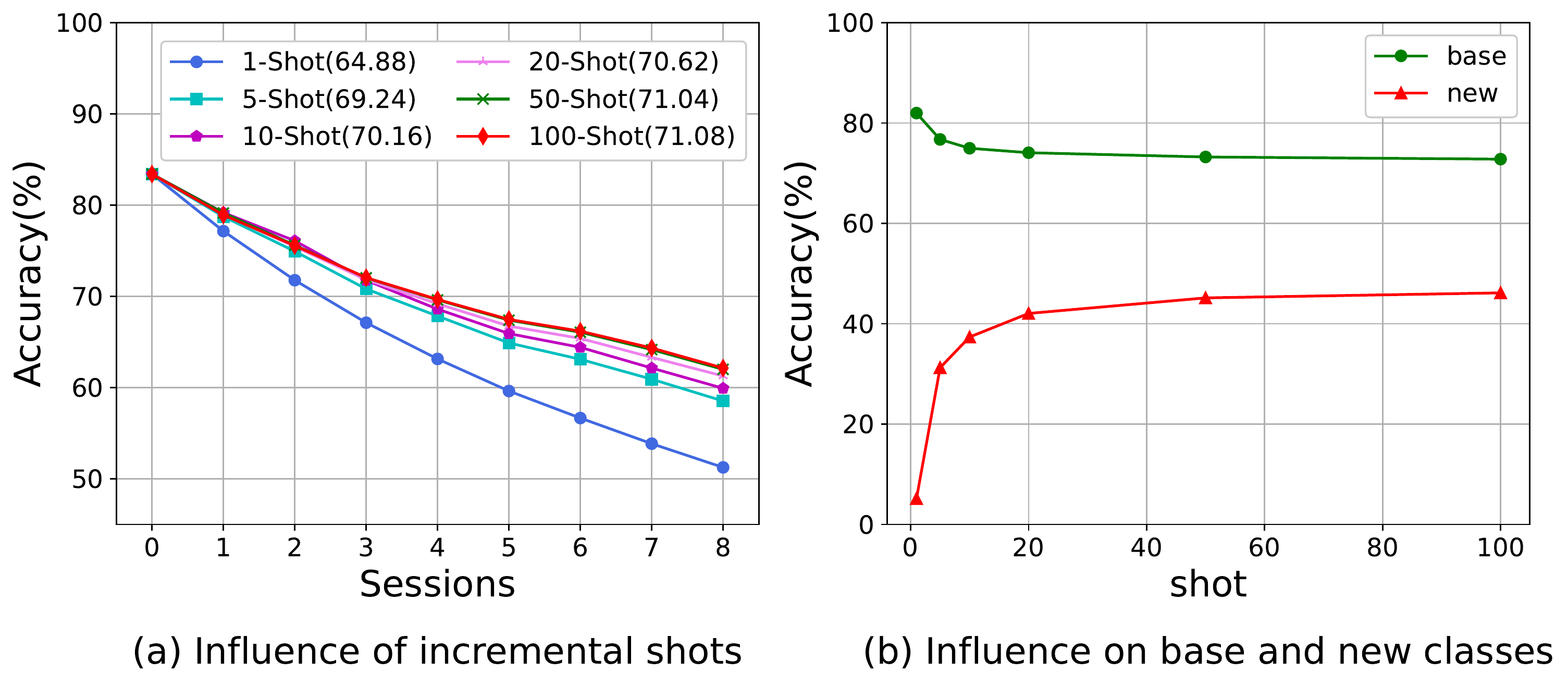}
\caption{Influence of incremental shots on (a) each session, and (b) base and new classes, where values shown in the bracket of the legend is the average accuracy across all sessions.
\label{fig:inc_shot}}
\end{figure}

\vspace{0.2cm}
\subsubsection{The sensitivity of the number of incremental shots.}
To explore the influence of the number of shots in inference stage, we change the number of shots among $\left\{1, 5, 10, 20, 50, 100\right\}$.
As can be seen from Figure \ref{fig:inc_shot}(a), increasing the number of shots from 1 to 5, the performance on incremental sessions is improved by a large margin.
However, such improvement gets small when we continually increase the number of shots.
The main reasons step from that increasing the number of shots from 1 to 5 slightly drops the model's performance on old classes but improves the model's performance on new classes by a large margin as shown in Figure \ref{fig:inc_shot} (b), and increasing the number of shots from 5 to a larger value seems to have a slight influence on old and new classes.

\vspace{0.2cm}
\subsubsection{Further analysis of RESA strategy}
To further verify the effectiveness of our RESA strategy, we compare RESA with the previously proposed knowledge transfer learning strategies RESS~\cite{Zhu_2021_CVPR}, PIL~\cite{Zhang_2021_CVPR}, and meta-learning~\cite{Chi_2022_CVPR}.
As we can see from Table \ref{tab:resa}, compared to the performance given by RESS (row1), PIL (row2), or meta-learning (row3), our proposed RESA (row 6) achieves the highest accuracy on each session on CUB200.
The results demonstrate that optimizing the model from the global perspective to coincide with the learning objective of FSCIL and the local perspective to improve the model's plasticity is a more effective strategy than global task-focused RESS, local task-focused PIL, and sequential local task-focused meta-learning.
Furthermore, we also try different strategies to synthesize the data.
We can see that using rotate \cite{Zhang_2021_CVPR} (row 6) can help our proposed method achieve better performance than the beneficial brought by using MixUp \cite{zhang2017mixup} (row 4) or CutMix \cite{yun2019cutmix} (row 5).
The results may show that rotate \cite{Zhang_2021_CVPR} is a more effective strategy than MixUp \cite{zhang2017mixup} and CutMix \cite{yun2019cutmix}.

\begin{figure*}[ht]
\centering
\includegraphics[width=1.9\columnwidth]{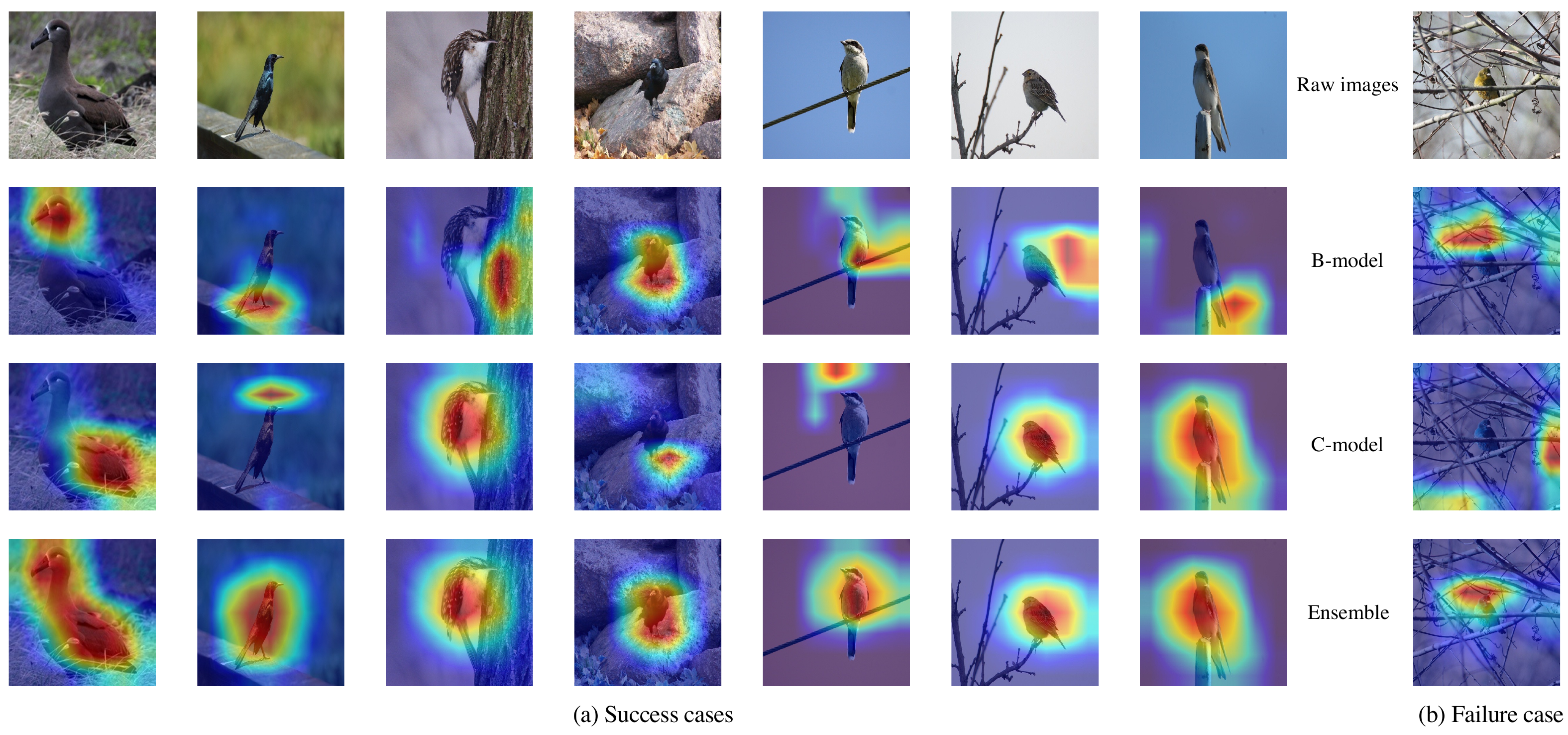}
\caption{Visualization with GradCAM on CUB200, where four old (column 1-4) and new classes (column 5-8) are selected, and images with different sizes are resized to the same size for the convenience of display. Our proposed method can well transfer the knowledge learned from the base session to the incremental sessions in most cases.
\label{fig:cam}}
\end{figure*}

\vspace{0.2cm}
\subsubsection{Visual explanation for knowledge transfer.}
To further give a visual explanation for our proposed method's knowledge transfer ability, we use GradCAM \cite{selvaraju2017grad} to visualize relevant results.
As shown in Figure \ref{fig:cam}, as long as the region of interest captured by the base model or the complementary model can relate to the target, our proposed method can better capture the target region (column 1-4). 
When processing new coming classes, we can see a similar phenomenon (column 5-7).
The results demonstrate that our proposed method can transfer the ability learned from the base session to the incremental sessions well.
However, if the region of interest captured by the base model and the complementary model do not involve the target, our proposed method fails (column 8).

\section{Conclusion}\label{sec:conclusion}
In this paper, we solve the few-shot class-incremental learning from two aspects, the effective knowledge transfer from the base session to the incremental sessions, and the design of robust metrics for incremental sessions.
For the first aspect, we propose a random episode sampling and selection strategy that mimics the real incremental setting and constructs pseudo incremental tasks from the global and local perspectives.
For the second aspect, we propose a simple yet effective method that utilizes complementary model with a squared euclidean-distance classifier as the auxiliary module, which couples with the widely used cosine classifier to perform incremental relation measuring.
Extensive experiments  on \textit{mini}ImageNet, CIFAR100, and CUB200 demonstrate that the  effectiveness of our proposed method named KT-RCNet compared with previous methods.

{\small
\bibliographystyle{unsrt}
\bibliography{main}
}

\end{document}